\documentclass[cameraready]{Interspeech}

\makeatletter

\renewcommand{\fnum@figure}{Fig. \thefigure}

\newcommand{\brq}{\texttt{BEST-RQ}}

\newcommand{\inlinesub}[1]{\vspace{0.2\baselineskip}\noindent\textbf{#1}}

\newcommand{\meta}{\texttt{Fair-speech}}

\newcommand{\sonos}{\texttt{Sonos}}
\usepackage{amsmath}
\usepackage{amssymb}
\usepackage{enumitem}
\usepackage{pifont}
\newcommand{\cmark}{\ding{51}}%
\newcommand{\xmark}{\ding{55}}%
\usepackage{makecell}

\newcommand{\wtvt}{\texttt{W2V2}}

\newcommand{\wavlm}{\texttt{WavLM}}
\newcommand{\whisper}{\texttt{Whisper}}

\newcommand{\largem}{\texttt{large}}

\usepackage{placeins}

\title{Where Do Self-Supervised Speech Models Become Unfair?}

\author[affiliation={1,2}, orcid=0009-0005-9771-4994, correspondingauthor]{Felix}{Herron}
\author[affiliation={2,3}]{Maja}{Hjuler} 
\author[affiliation={2}]{Solange}{Rossato}
\author[affiliation={1}]{Alexandre}{Allauzen} 
\author[affiliation={2}]{François}{Portet} 

\address{
    $^1$ MILES Team, LAMSADE, Université Paris Dauphine-PSL\\
    $^2$ GETALP Team, LIG, Université Grenoble Alpes \\
    $^3$ Queensland University of Technology
}

\email{felix.herron@univ-grenoble-alpes.fr}


\usepackage{comment}

\begin{document}

\maketitle

\begin{abstract}
    Speech encoder models are known to model members of some speaker groups (SGs) better than others. However, there has been little work in establishing why this occurs on a technological level. To our knowledge, we present the first layerwise fairness analysis of pretrained self-supervised speech encoder models (S3Ms), probing each embedding layer for speaker identification (SID) automatic speech recognition (ASR). We find S3Ms produce embeddings biased against certain SGs for both tasks, starting at the \textit{very first} latent layers. Furthermore, we find opposite patterns of layerwise bias for SID vs ASR for all models in our study: SID bias is \textit{minimized} in layers that \textit{minimize} overall SID error; on the other hand, ASR bias is \textit{maximized} in layers that \textit{minimize} overall ASR error. The inverse bias/error relationship for ASR is unaffected when probing S3Ms that are finetuned for ASR, suggesting SG-level bias is established during pretraining and is difficult to remove.
\end{abstract}

\section{Introduction}



Self-supervised speech processing models (S3Ms) have been repeatedly shown to perform better for certain Speaker Groups (SGs) than others on tasks such as automatic speech recognition (ASR) \cite{asr_fair_benchmark,fairspeech_bad_methodology_asr,bhattacharjeeFairnessAutomaticSpeech2025} and speaker identification (SID) \cite{big_thorough_bias_survey}. Despite many attempts at reducing this fairness gap \cite{adversarial_doesnt_work_for_unseen,survey_on_accents_det_dat,hend_gender,solange_balanced,sid_fairness,fairspeech_bad_methodology_asr}, none has come near to closing it; one potential explanation for this intractability is our limited collective understanding of how bias propagates through S3Ms. In particular, there is no consensus as to where - starting at which layer, and to what magnitude - S3Ms diverge in their ability to model speech of different SGs.

We tackle this question by performing a layerwise analysis of the quality of S3Ms' latent embeddings with respect to both ASR and SID. At each layer, we measure both the overall task error rate and the relative error rate for each SG, in order to quantify bias in layerwise embedding quality (difference in by-SG error rate for each task). First, we investigate pretrained S3Ms in order to benchmark their behavior prior to finetuning; then, we turn to finetuned models to see whether the biases already baked into pretrained models are amplified or smoothed out. Formally, we pose three principal research questions.

\begin{enumerate}[label=\bf RQ \arabic*,itemindent=*]
    \item Starting at which S3M layers do we observe performance bias for each SG on each task? \label{rq:where}
    \item What is the layerwise relationship between overall performance for both tasks and bias against certain SGs? \label{rq:fair_vs_perf}
    \item How does ASR finetuning (with or without a fairness-enhancing algorithm) change behavior observed in \ref{rq:fair_vs_perf}? \label{rq:ft}
\end{enumerate}





Regarding \ref{rq:where}, we fine that pretrained S3Ms produce higher quality embeddings for certain SGs starting at the very first self-attention layer of all S3Ms, for both SID and ASR. However, we find that the SGs which experience greatest bias are not the same: non-native speakers and children for ASR, and children and women for SID. (This aligns with previous studies on SG-level bias in ASR and SID \cite{sonos,bias_mitigation_sid}). Regarding \ref{rq:fair_vs_perf}, we find \textbf{opposite behaviors} in the relationship between bias and performance for ASR and SID: SID performance peaks at the earliest S3M layers, while SG-level bias is at its lowest; as overall SID performance degrades at each subsequent layer, bias increases. Thus, we find an \textit{inverse relationship} between SID performance and bias against certain SGs. On the other hand, both ASR performance \textit{and} bias increase at each subsequent layer until ASR performance is maximized. Thus, we find a \textit{direct} relationship between ASR performance and bias. Regarding \ref{rq:ft}, we find that while finetuning for ASR increases overall performance, it has little effect on the SG-level bias, even when finetuned with the fairness-enhancing domain enhancing/adversarial training algorithm. This supports the hypothesis that SG-level bias is baked into S3Ms during pretraining, and it is hard to remove during finetuning.


\section{Background}

\textbf{Self-supervised learning (SSL)} Two popular techniques for training self-attention based speech processing models are 1) SSL + task-specific finetuning, and 2) task-specific end-to-end training \cite{towards}. The first method trains on a large unlabeled audio corpus using SSL objectives like contrastive loss, followed by adaptation to specific downstream tasks with limited labeled data \cite{wav2vec2,hubert}. The second method trains exclusively on labeled data for particular tasks like ASR or SID using task-specific losses \cite{whisper}. Principal advantages of SSL over end-to-end training include a) reduced dependence on labeled data, and b) easy adaptation to heterogeneous downstream tasks. Furthermore, while S3Ms are typically finetuned for downstream tasks, embeddings learned by pretrained S3Ms \textit{already} contain information pertinent to multitudinous tasks \cite{wavlm,superb}. While end-to-end ASR models achieve superior performance \cite{whisper}, the best-performing ones have been found to be less fair than their lower-performing counterparts \cite{asr_fair_benchmark} and finetuned S3Ms \cite{towards}.



\textbf{Existing techniques for fairer training} There is a breadth of study on fairer training for ASR; however, none has yet been shown to eliminate the performance gap between SGs. This research can be categorized into \textit{model-based} and \textit{data-based} approaches. \textit{Model-based} techniques adapt the finetuning algorithm or model architecture to better suit disadvantaged SGs. For example, enhancing/adversarial training (DET/DAT) emphasizes/blinds models to speaker identity in middle/later layers respectively \cite{adversarial_doesnt_work_for_unseen,survey_on_accents_det_dat,dat_english_accents,accent_embeddings_dialect}. Another well-studied approach involves steering vectors containing auxiliary utterance information, which could be comprised of speaker \cite{survey_on_accents_det_dat} or accent information \cite{accent_embeddings_dialect}, or unsupervised information pertaining to predicted ASR failure \cite{amazon_kmeans}. \textit{Data-based} approaches involve balancing training data to avoid underexposure of typically under-represented SGs \cite{hend_gender,hend_gender_emnlp,laura_kmeans,commonvoice_data_based,adversarial_and_enhancing}.



Papers studying the behavior of S3Ms' treatment of differing speaker groups typically evaluate them after finetuning \cite{mtl_dat_comp,bhattacharjeeFairnessAutomaticSpeech2025,masson_phoneme_similarity_tts,fairspeech_bad_methodology_asr,asr_fair_benchmark,dutch_quantifying}. To our knowledge, no studies have evaluated SG-level bias in pretrained-only models for ASR or SID bias\footnote{One paper that tangentially touch upon this idea shows that pretrained models encode human-like sociological biases, influencing its performance on speech emotion recognition \cite{slaughterPretrainedSpeechProcessing2023}.}. This gap in the literature is a critical shortcoming for several reasons. First, pretrained S3M embeddings are often used out of the box \cite{wavlm,yuxuan_emotion_pretrained}. Second, by better understanding where and how bias appear in pretrained models, we can a) more precisely develop fairer pretraining techniques and b) ascertain the utility of fairer finetuning. If pretrained S3Ms are already biased against certain SGs, erasing bias via fairer finetuning might be an unreasonable objective.

\section{Methodology}


\subsection{Lightweight SID and ASR decoders at every layer}
\label{sec:probes}



We seek to measure each layer's ability to model both SID and ASR, both overall \textit{and} relatively for each SG. To measure this, we train lightweight decoders for both ASR and SID based on embeddings from each layer of each encoder model. Following the SUPERB framework, our decoders are comprised of a single linear layer\footnote{Whereas SUPERB aggregates representations from all layers together to form one vector, we learn a different decoder at each layer.}, thereby reducing learnable bias during finetuning as much as possible \cite{superb}. This simple decoder architecture will inevitably deflate overall performance compared with SOTA deep decoders \cite{joint_encoder_decoder}; however, downstream performance maximization is not the point of this study. To the contrary, the fact that our decoders are so light means that they will be relatively immune to learning their own biases during training; thus any bias we observe necessarily comes from the pretrained S3M\footnote{One potential weakness of our methodology is that a more powerful decoder might be able to correct for some SG-level bias that a linear layer cannot. However, given previous results showing prevalent SG-level bias even in the most powerful models, we suspect that a deep attention-based decoder would not remove the effects we observe.}. Inasmuch as they are linear transformations based on frozen latent representations, we will from hereon out refer to our linear SID/ASR decoders as ``probes''. We replicate each experiment five times for robustness, and train for 5k/30k steps (for SID/ASR respectively) using dynamic batches of maximum 3 minutes in length. We based our code on Speechbrain's ASR/SID recipes for CommonVoice respectively \cite{speechbrain}.

We train and evaluate our probes using the Sonos Voice Control Bias Assessment Dataset \cite{sonos} (\sonos{}) as well as Meta's Fair-speech corpus (\meta{}) \cite{meta_fair}. Both corpora were designed for evaluation of SG-level bias in ASR models and are annotated with SG-level metadata for several demographic variables. \sonos{} includes speaker ID (pseudo-id), gender, regional dialect (for USA speakers), age group, and native language (is\_native); \meta{} includes gender, age,  ethnicity, native language (is\_native) and socioeconomic background. \sonos{} contains 1038 speakers for a total of 166 hours; \meta{} contains 593 speakers over 56 hours.

We train \textbf{SID probes} on \sonos{} (\meta{} doesn't contain speaker IDs) using the entire set of speakers, ignoring the predefined train/valid/test ASR partitions. For each replication, we randomly sample 15 utterances for each speaker, creating our own 80/20 train/test splits. For \textbf{ASR probes} on \sonos{}, we train using the provided ``train'' split (the \sonos{} splits are balanced along demographic characteristics \cite{sonos}). \meta{} doesn't contain speaker IDs, so it cannot be used for training out of the box; to circumvent this, we estimated Speaker IDs by clustering on utterance embeddings extracted from XLS-R. We use these clustered pseudo-IDs to define our own ``likely speaker-disjoint" train/test split.




\subsection{Measuring overall error and SG-level bias}

Formally, we train probes on a task $T$ (SID or ASR) based on latent embeddings from pretrained S3M $M$, sampled at a given layer $\ell$, and evaluate them on an unseen test set $D$. For the sake of simplicity, we define the triple $T^+ := (T, M, \ell)$ as the task, model, and layer on which a probe was trained. Let us denote $\text{Err}: (U; T^+)  \to \mathbb{R}$ as an utterance-wise error metric corresponding to probe configuration $T^+$, and $\text{Err}_{avg}(D,SG; T^+) = \frac{1}{D \cap SG}\sum\limits_{u \in D \cap SG} \text{Err}(u; T^+)$ as average error rate for any SG. We use classification error rate (SIDER), and WER as the utterance-wise error metric $\text{Err}(U, T^+)$ for SID and ASR respectively.

We define the relative error rate for a SG with respect to the general population. This is a commonly used metric in the fairness literature \cite{hend_gender_emnlp,towards,fairness_definitions} \footnote{Other studies, such as \cite{fairspeech_bad_methodology_asr} or \cite{gender_perf_gaps_multilingual_pebbles} measure a normalized relative error gap. We eschewed this in order to compare bias with overall error; $\frac{\vec{a}}{\vec{b}}, \vec{b}$ are dependent, thus ruining any statistical test comparing the two.}. Ideally this would be as close to zero as possible for every $SG$, representing statistical parity in the fairness taxonomy according to \cite{fairness_definitions}:

\begin{align}
    &\hspace{-0.5cm}\text{Err}_{rel}(D, SG; T^+) :=\text{Err}_{avg}(D,SG; T^+)-\text{Err}_{avg}(D; T^+) \hspace{-0.1cm} \label{eq:rel_err}
\end{align}

\noindent We can harness this metric to subsequently measure the overall bias for a demographic variable $DV$ (e.g. gender) of a $T^+$ triple with respect to a test set $D$ by calculating the mean absolute relative error for each $SG \in DV$:

\begin{align}
    &\hspace{-0.2cm}\text{DV-bias}(D, DV; T^+) := \frac{1}{|DV|} \sum\limits_{SG \in DV}\hspace{-0.15cm}|\text{Err}_{rel} (D, SG; T^+)|  \label{eq:dv_bias}
\end{align}

\noindent We choose DV-bias over the related min-max measure \cite{minimax_rawls} as it takes other SGs into account apart from the extrema.


\subsection{S3Ms included in our study}

\begin{table}[th!]
    \caption{Overview of Pretrained S3Ms (and \whisper{})} 
    \scriptsize 

  \centering
  \centerline{
  \addtolength{\tabcolsep}{-0.3em}

\begin{tabular}{lllll}
    \toprule
      model name & \makecell[l]{num.\\params} & \makecell[l]{pretrain\\hours} & \makecell[l]{multi-\\lingual} & architecture \\\midrule
     \wavlm{}-base+ \cite{wavlm} & 100M & 94k & \xmark & transformer\\
     \wavlm{}-lg \cite{wavlm} & 300M & 94k & \xmark & transformer\\\midrule
     \brq{}-lg-ll \cite{brq,bestrq_ryan} & 300M & 60k & \xmark & conformer\\\midrule
     \wtvt{}-lg-ls \cite{wav2vec2} & 300M & 960 & \xmark & transformer\\
     \wtvt{}-lg-lv \cite{wav2vec2} & 300M & 60k & \xmark & transformer\\
     \wtvt{}-lg-xlsr-53 (XLS-R) \cite{w2v2-xlsr}& 300M & 436K & \cmark & transformer\\
     \wtvt{}-FR-7K-lg \cite{lebenchmark_2} & 300M & 7k & \xmark & transformer\\\midrule
     \whisper{}-med \cite{whisper} & 300M & 680k & \cmark & \makecell[l]{transformer (E2E training)} \\
    \midrule
    \bottomrule
    \label{tab:models}
    \end{tabular}
    }
    \caption*{Note several abbreviations: \wtvt{} denotes \texttt{wav2vec 2.0}; lv denotes LibriVox \cite{wav2vec2}, ll denotes LibriLight \cite{librilight}, ls denotes LibriSpeech \cite{librispeech}; suffix ``lg'' denotes large models size ($\approx$ 300M parameters). \whisper{}-medium has a similar number of parameters as \largem{} S3Ms. XLS-R was pretrained on 53 languages (including English); \wtvt{}-FR-7K-lg only on French.  \vspace{-0.5cm}}

     \label{tab:asr_macro_total}
\end{table}

\inlinesub{Pretrained S3Ms} We use several classes of SOTA pretrained S3Ms as a starting point for our study, varying in size, architecture, and pretraining data, as depicted in Table \ref{tab:models}. This allows us to determine whether these aspects influence evolution in layerwise bias. We use one multilingual S3M and another pretrained on French in order to ascertain whether unequal SG-level performance is due to the biases inherent to unbalanced pretraining corpora \cite{solange_bias_women}. Additionally we compare the \whisper{}-medium encoder, which is not an S3M (it is trained end-to-end on transcription and translation), and the embeddings it produces are not meant to stand alone. However, \whisper{} achieves SOTA performance in ASR, thus it is informative to study its encoder.

\inlinesub{Finetuned models for ASR} To test whether finetuning can resolve the bias patterns identified in pretrained S3Ms, we train layerwise ASR probes on S3Ms finetuned for ASR using a subset of English CommonVoice 16.0 (5k frozen warmup then 25k unfrozen steps with constant $lr=1e-4$). As a representative fairness-enhancing baseline, we additionally experiment with domain-enhancing/adversarial training (DET/DAT). For the implementation, we follow \cite{adversarial_and_enhancing}, putting DET classifier on layers 5/10, and DAT classifier on 10/21 (for base/large configurations, respectively). We use the xvector architecture for our enhancing and adversarial classifiers \cite{xvectors}, motivated by \cite{herron_tsd}, which shows that DET + DAT using xvectors effectively removes SG information better than linear classifiers.

\begin{figure}[t!]
  \includegraphics[width=\linewidth]{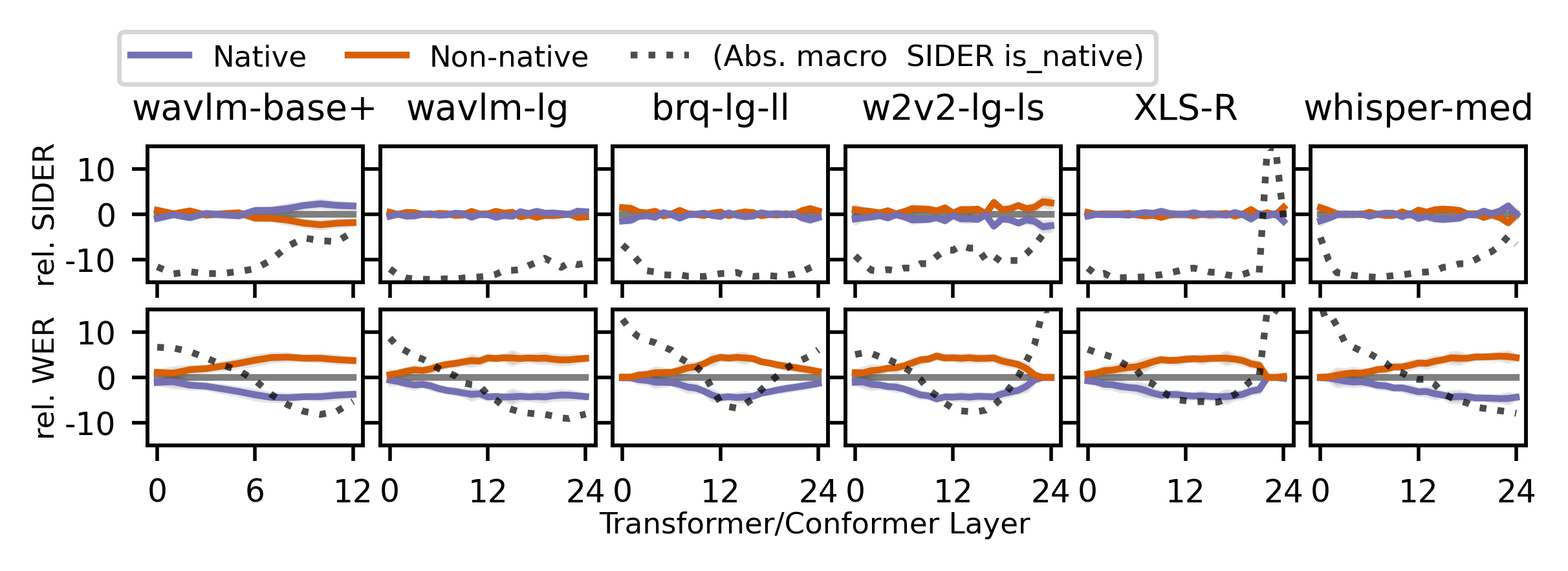}
  \caption{Layerwise evolution of relative error rate (see Eq. \ref{eq:rel_err}) for \textbf{native/non-native speakers} (solid lines) vs overall error (dotted). Relative error $>0$ implies an \textit{error rate} higher than average, i.e. worse performance. Non-native speakers have increasingly worse relative performance from layer to layer for ASR, though near-equal performance for SID. This is in contrast to by-age (see Fig. \ref{fig:rel_perf_age}).}
  \label{fig:rel_perf_is_native}
\end{figure}

\begin{figure}[t!]
  \includegraphics[width=\linewidth]{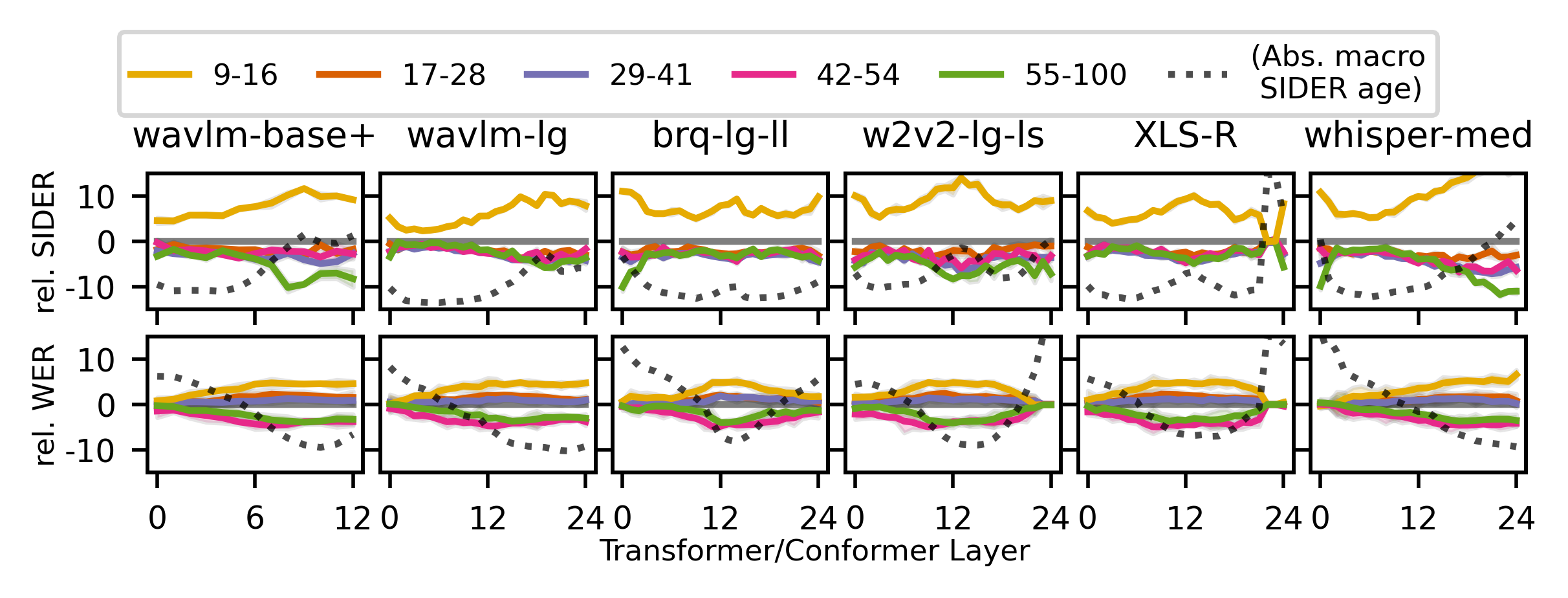}
  \caption{Layerwise evolution of relative error for different \textbf{ages} (solid lines) vs overall error (dotted). Children (9-16) are worst modeled while older adults (42+) are best modeled, for both SID and ASR. Like Fig. \ref{fig:rel_perf_is_native}, ASR bias grows as overall ASR error \textbf{shrinks}; unlike Fig. \ref{fig:rel_perf_is_native}, SID bias grows as overall SIDER \textbf{grows}.}
  \label{fig:rel_perf_age}
  
\end{figure}

\begin{figure}[b!]
  \includegraphics[width=\linewidth]{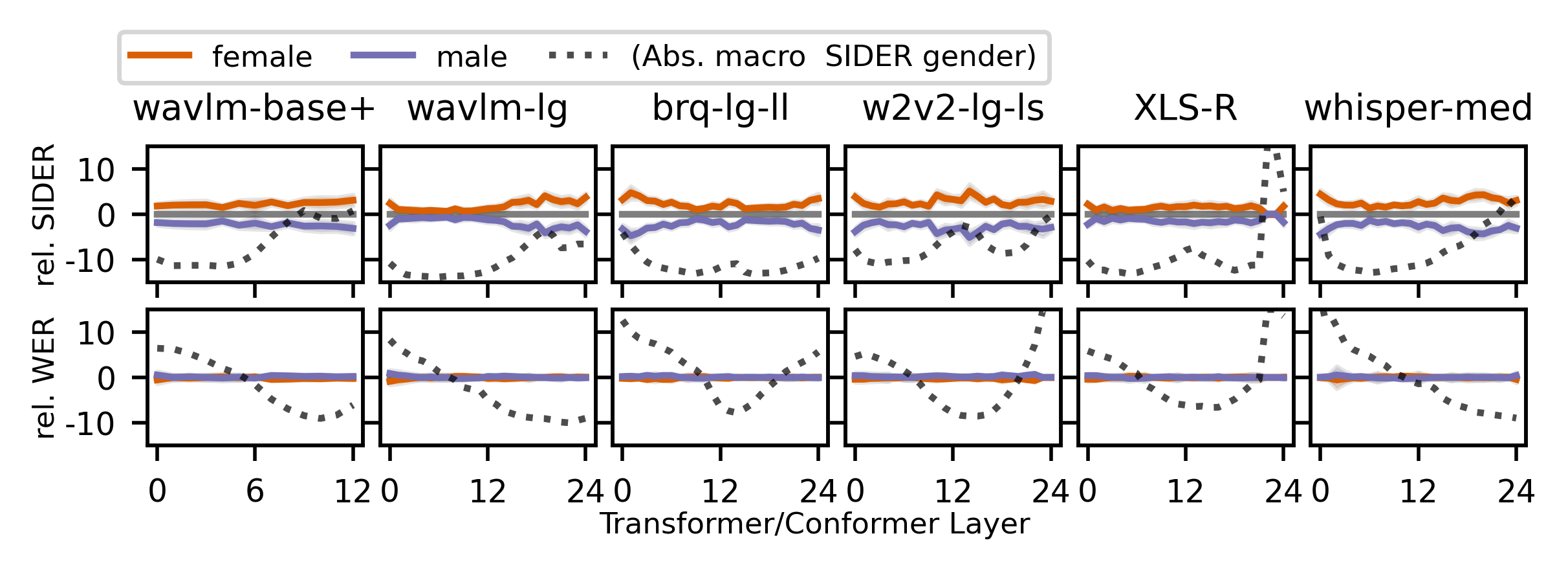}
  \caption{Layerwise evolution of relative \textbf{gender} error (solid lines) vs overall error (dotted). Women experience worse SID performance at most layers; there is negligible layerwise gender-level bias evolution for ASR.}
  \label{fig:rel_perf_gender}
\end{figure}

\begin{figure}[b!]
    
  \includegraphics[width=\linewidth]{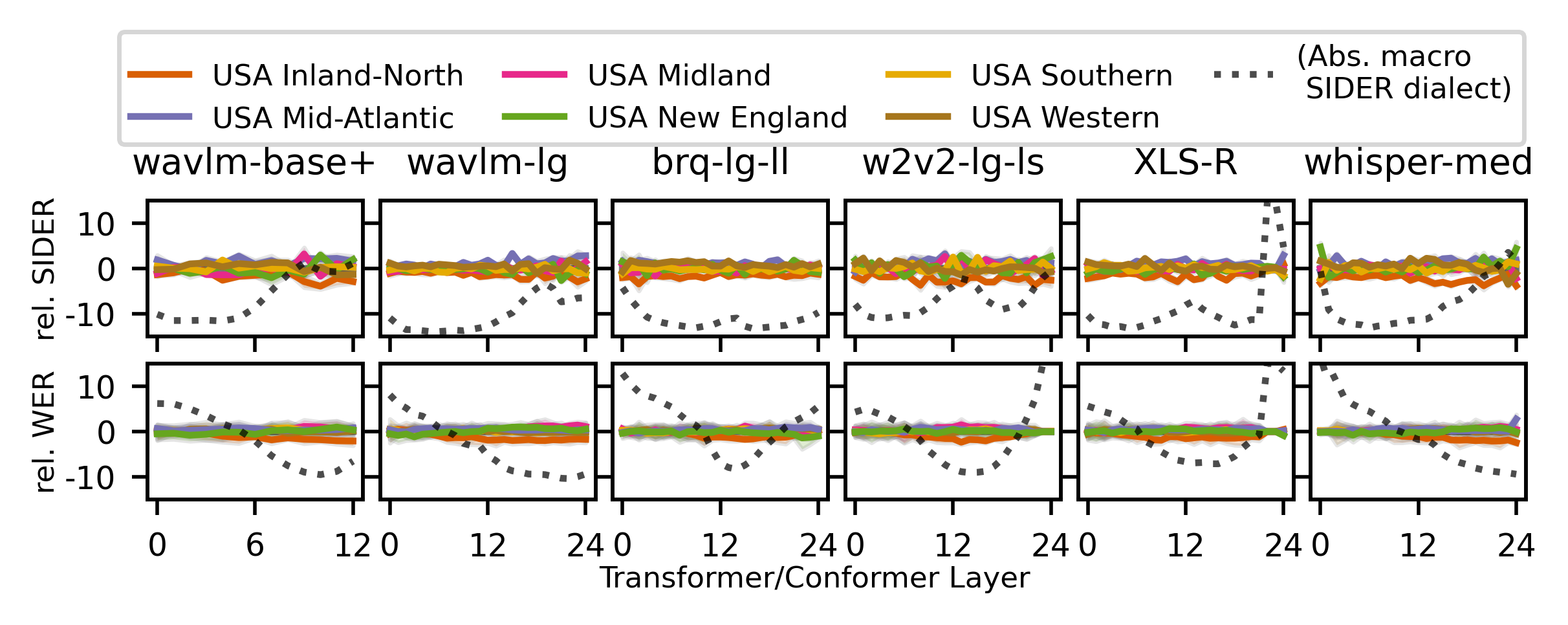}
  \caption{Layerwise evolution of relative \textbf{dialect} error (solid lines) vs overall error (dotted) for American native-English speakers.}
  \label{fig:rel_perf_dialect}
\end{figure}


\begin{figure*}[ht!]
  \begin{center}
    \includegraphics[width=\linewidth]{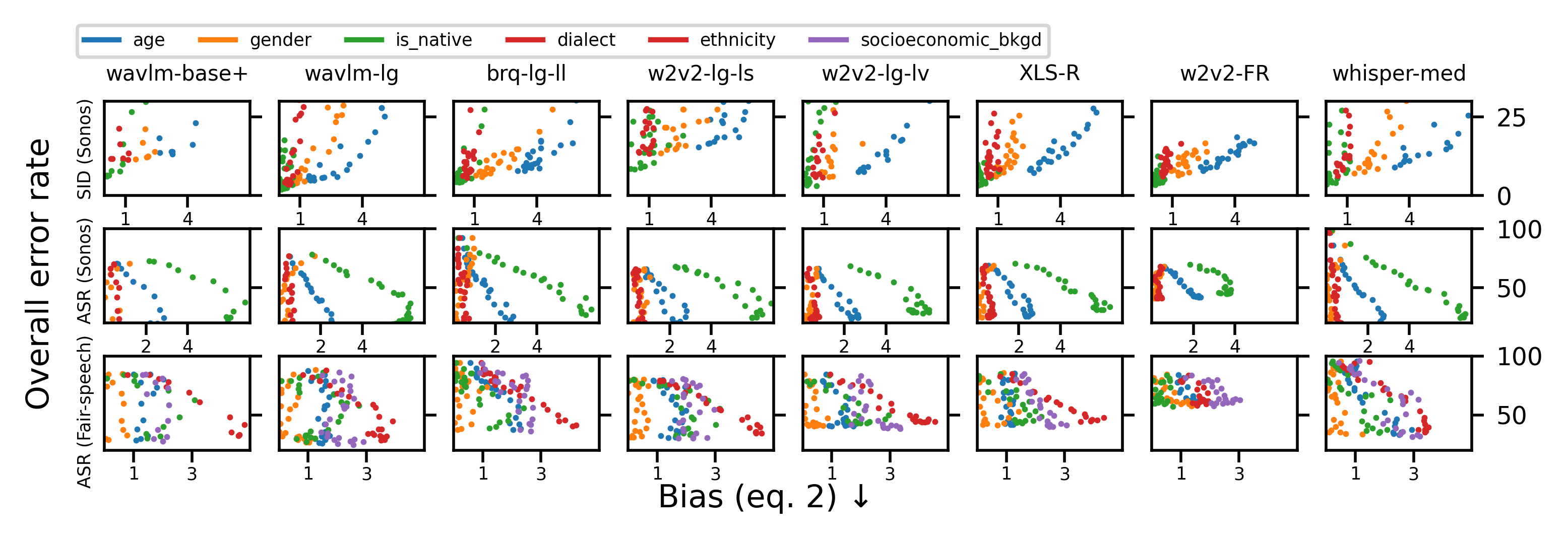}
  \end{center}
  \vspace{-0.4cm}
  \caption{Each dot represents one layer for each pretrained S3M, plotted according to its bias and overall error rate. Layers with low bias and low error will be at the bottom left of each plot; layers with high bias and low error will be at the bottom right of each plot. For SID, layers with low overall error also have low SG-level bias; for ASR, layers with low overall error have high SG-level bias.}
  \label{fig:fairness_vs_perf}
\end{figure*}

\section{Probing results}

Due to space constraints, we focus on a representative subset of S3Ms tested on \sonos{}; we integrate all S3Ms and \meta{} in Fig. \ref{fig:fairness_vs_perf}.

\subsection{Layerwise SG-level bias for SID and ASR}

Figs. \ref{fig:rel_perf_is_native}-\ref{fig:rel_perf_dialect}  depict relative error rates in \sonos{} for is\_native, age, gender, and dialect respectively, for both SID and ASR probes. We obtain overall layerwise error rates similar to previous studies, i.e. SID performance peaks in earlier layers while ASR performance peaks in later layers  \cite{wavlm,adversarial_and_enhancing}\footnote{We find, as in previous studies, that final layers of \wtvt{} models abruptly shed ASR-relevant information \cite{toyota_layerwise_interpretability,herron_tsd}.}. All models we tested exhibit roughly the same relationship between SID/ASR fairness and bias, as Fig. \ref{fig:fairness_vs_perf} illustrates. Surprisingly, multilingual and non-English S3Ms display the same patterns of bias/performance ratio towards non-native speakers as S3Ms pretrained only on English.



    

\inlinesub{SID error rate vs SG-level bias} In the earliest S3M layers, where SID performance is maximized, SID bias is at its lowest for every model and every SG. It is only at layers where overall SID performance begins to degrade that bias against certain SGs (women, children) begins to grow. In other words, layers that encode the most SID information do so nearly equally well for all demographic variables. This is evidence of a \textbf{direct relationship} between SID error rate and SID bias - as SID error rate goes down, SG-level bias goes down too\footnote{We find positive statistically significant $p < 0.05$ Pearson's $r$ between SIDER and SG-level SID bias for all S3Ms on age and gender.}. This trend is directly illustrated in the first row of Fig. \ref{fig:fairness_vs_perf}. We note that layers with poor SID performance are unlikely to be used in any SID-related downstream tasks, so their unfair performance at those layers is practically irrelevant. Thus, we focus more on ASR for the remainder of the study.

\inlinesub{ASR error rate vs fairness} We observe the opposite trend in ASR probes. The earliest model layers, where ASR performance is lowest, tend to deliver the embeddings with the least SG-level bias; as WER decreases, SG-level bias \textit{increases}. For some models (\whisper{}-med, \brq{}-lg-ll) bias continues to grow until the layer where WER performance is maximized, while for the remaining models bias bottoms out and even diminishes slightly in the layers directly preceding maximum ASR performance. However as a general trend, we observe that layers which maximize ASR performance also deliver near-maximal SG-level bias\footnote{We find negative statistically significant $p < 0.05$ Pearson's $r$ between WER and SG-level ASR bias for all S3Ms on age, is\_native, and dialect (except dialect for \wtvt{}-FR) on \sonos{}. For \meta{} we find $p < 0.05$ on ethnicity for all S3Ms; is\_native and age for all but the \wavlm{} models and \wtvt{}-FR.}. This is evidence of an \textbf{inverse relationship} between ASR error rate and ASR bias, illustrated in the second and third rows of Fig. \ref{fig:fairness_vs_perf}.

\begin{figure}[h!]
  \includegraphics[width=\linewidth]{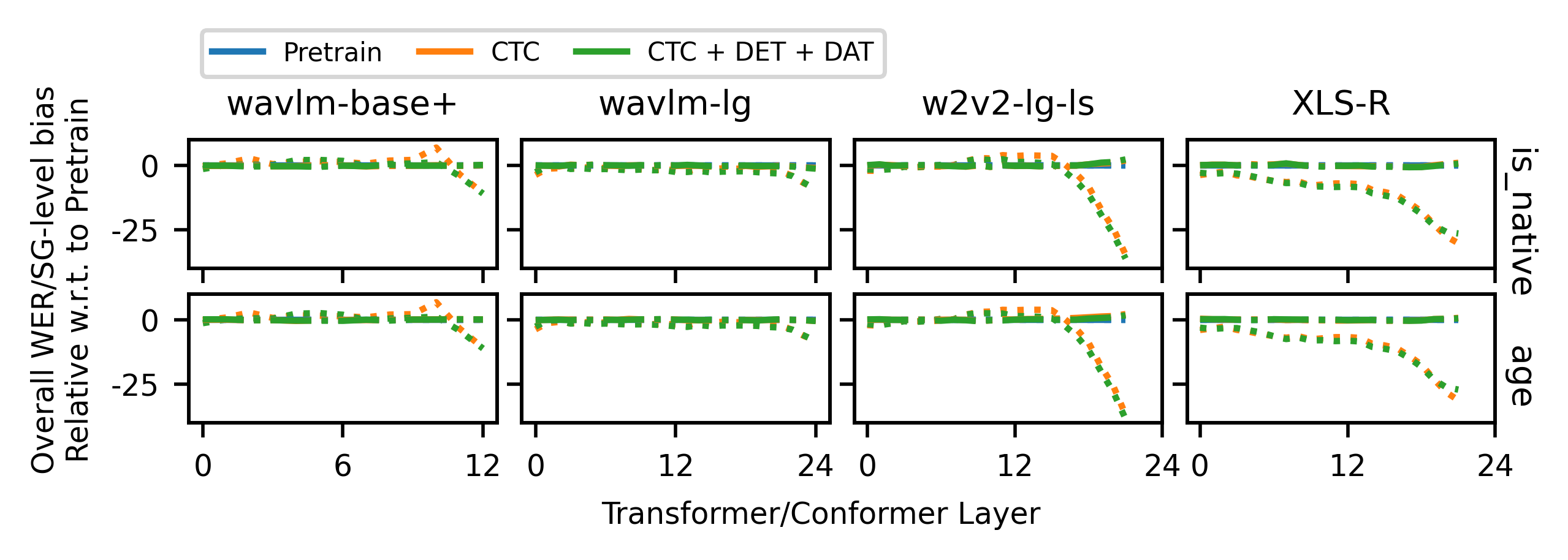}
  \caption{Relative overall WER (dots) and SG-level bias (Eq. \ref{eq:dv_bias}, dot-dash) for ASR finetuned models relative to pretrained S3Ms on \sonos{} (see Eq. \ref{eq:rel_err}). Values $<0$ means lower WER (dots) or less bias (dot-dash) than the pretrained S3M respectively.}
  \label{fig:with_ft}
\end{figure}

\subsection{Effect of finetuning}

We repeat our experiments on S3Ms that have been finetuned for ASR, both using the vanilla CTC algorithm as well as using the fairness-enhancing CTC + DET + DAT \cite{adversarial_doesnt_work_for_unseen,adversarial_and_enhancing}. Fig. \ref{fig:with_ft} shows the relative overall error and SG-level bias of finetuned S3Ms with respect to pretrained S3Ms. We find, unsurprisingly, that ASR finetuning improves overall performance, particularly in later layers. However, we find negligible effect on bias - both finetuning variants retain similar amounts of SG-level bias as the retrained model. Hence, finetuning is clearly not a sufficient answer to the bias patterns baked into S3Ms during pretraining.

\section{Conclusion and Future Work}

In this paper we probed speech encoder models at each layer for two complementary downstream tasks, ASR and SID. Regarding \ref{rq:where}, we find the very first layers of each S3M produce embeddings biased against certain SGs for both tasks. This implies that unfairness is rooted deeply within conventionally pretrained S3Ms. Regarding \ref{rq:fair_vs_perf}, we find that overall SID error and SID SG-level bias vary directly layerwise, while overall ASR error and ASR SG-level bias vary inversely. These findings suggest that S3Ms learn local/ASR-pertinent features at the behest of certain SGs and at the expense of others, while they learn utterance-level/SID-pertinent features equally well for all SGs. Finally, regarding \ref{rq:ft}, we find that ASR finetuning doesn't change the aforementioned layerwise bias/performance trends, even using domain-enhancing/adversarial training. This provides an explanation for the modest results that have been observed in practice for DET + DAT in bias mitigation \cite{using_dat_vc_dutch,adversarial_doesnt_work_for_unseen,dat_english_accents}. To the contrary, it motivates further study into fairer pretraining techniques, so that S3Ms finetuning doesn't have to correct for bias as well.

Furthermore, we find similar layerwise SG-level bias patterns in S3Ms trained on small homogeneous corpora, on diverse English data, multilingual and even exclusively non-English data. We conclude that there are likely inherent challenges in modeling certain SGs endemic to the dataset. Even the earliest layers, which are overall entirely ineffective at ASR, are slightly worse for the same SGs which experience the greatest discrimination in the best ASR layers.

With this in mind, we encourage further work into explaining the causal mechanisms driving SG-level bias in S3M embeddings. In particular, given biases inherent in SG-level metadata in common fairness evaluation corpora \cite{meta_fair,artie}, we propose focusing on attributes of individuals and individual utterances rather than SG labels. If we isolate specific properties of speakers with worse downstream task performance, such as dialect density measure \cite{black_white_princeville} or cadence \cite{mengDontSpeakToo2022}, we could better understand the layerwise bias patterns uncovered in this study. 




\FloatBarrier
\bibliographystyle{IEEEtran}
\bibliography{these}

\end{document}